# High Speed Cognitive Domain Ontologies for Asset Allocation Using Loihi Spiking Neurons

Chris Yakopcic[1*], Nayim Rahman[1], Tanvir Atahary[1], Tarek M. Taha[1], Alex Beigh[2], and Scott Douglass[3]
[1]*Dept. Of Electrical and Computer Engineering*, University of Dayton, Dayton, OH, USA
[2]*University of Dayton Research Institute*, Dayton, OH, USA
[3]*Human Effectiveness Directorate, Air Force Research Laboratory, Wright Patterson Air Force Base*, OH, USA
*cyakopcic1@udayton.edu

*Abstract*—Cognitive agents are typically utilized in autonomous systems for automated decision making. These systems interact at real time with their environment and are generally heavily power constrained. Thus, there is a strong need for a real time agent running on a low power platform. The agent examined is the Cognitively Enhanced Complex Event Processing (CECEP) architecture. This is an autonomous decision support tool that reasons like humans and enables enhanced agent-based decision-making. It has applications in a large variety of domains including autonomous systems, operations research, intelligence analysis, and data mining. One of the key components of CECEP is the mining of knowledge from a repository described as a Cognitive Domain Ontology (CDO). One problem that is often tasked to CDOs is asset allocation. Given the number of possible solutions in this allocation problem, determining the optimal solution via CDO can be very time consuming. In this work we show that a grid of isolated spiking neurons is capable of generating solutions to this problem very quickly, although some degree of approximation is required to achieve the speedup. However, the approximate spiking approach presented in this work was able to complete all allocation simulations with greater than 99.9% accuracy. To show the feasibility of low power implementation, this algorithm was executed using the Intel Loihi manycore neuromorphic processor. Given the vast increase in speed (greater than 1000 times in larger allocation problems), as well as the reduction in computational requirements, the presented algorithm is ideal for moving asset allocation to low power, portable, embedded hardware.

## I. INTRODUCTION

Autonomous systems are being increasingly utilized in a variety of domains, including both mobile systems (UAVs, cars, and robots) as well as planning systems. In these systems, cognitive agents are utilized for autonomous decision making. Multiple cognitive architectures have been developed over the years [1-6], among which SOAR [3] and ACT-R [4-6] are two of the most widely explored. Cognitive scientists have combined complex event processing and cognitive modeling in a cognitively enhanced complex event processing (CECEP) architecture [7-12]. The high-performance complex event processing technology at the core of the CECEP architecture distinguishes it from other cognitive modeling frameworks and architectures. This enables it to process very large knowledge bases and thus enhances its cognitive capabilities. It is well suited to the challenges of developing autonomous decision support tools that reason and learn like humans.

Rather than focusing on the general characteristics of the CECEP architecture, this paper focuses on the processing of domain knowledge in one event processing component of CECEP. This component, soaCDO, is a knowledge representation and mining application that allows a cognitive agent to store and exploit domain knowledge. Domain knowledge represented in a Cognitive Domain Ontology (CDO) should enable CECEP agents operating in military, civil, and commercial contexts to act autonomously or provide decision assistance in: 1) Operations Research, 2) Course of Action Analysis/Comparison, 3) Generic Planning, 4) Intelligence Analysis, 5) Forensic Analysis, and 6) Data Mining. Thus CDOs can store knowledge from a wide variety of application domains and process them with complex constraints. While complex CDOs increase the capabilities of decision agents, they are computationally expensive. Therefore, mapping them to novel, energy efficient hardware [13-19] may provide significant power advantages to autonomous systems [20-23].

In this paper, we present a spiking neural network implemented to alleviate the power requirements of complex CDOs. More specifically, in this work we propose an approximate solution to the asset allocation problem using a grid of spiking neurons. Given the number of possible solutions in this allocation problem, determining the optimal allocation via typical CDO decision methods can be very time consuming. This is because solving this problem involves an exhaustive search over a very large set of possible solutions. The proposed spiking approach is approximate, but orders of magnitude faster in comparison. In this work we carry out a number of allocation simulations and show that the spiking approach is capable of generating a solution that is more than 99.9% accurate in all cases, with a potential speedup greater than 1000×. In order to show that this system can be implemented on low power hardware, the proposed algorithm was executed on the Intel Loihi Spiking manycore neuromorphic processor [24-26]. The studied in this work were executed via remote login to a physical Loihi system.

While alternative algorithms for asset allocation have been proposed [27-31] that do not involve an exhaustive search [11-12], we are not aware of any spiking neural network solutions for asset allocation besides [23]. Furthermore, since the recent release of the Loihi processor, very few studies that have utilized the system have been published [24-26]. Therefore, this paper presents one of the first spiking algorithms for assessing cognitive domain knowledge, and also presents one of the first low power implementations of a cognitive domain ontology on specialized spiking hardware. With this work we aim to show the efficacy of spiking neural networks as applied to inference engines with extremely large solution spaces for drastic reductions in computation time.

The work in the following sections of this paper is based on previous research [20-23] that explores the reduction of CDO computation time and power consumption. Work in [20,21] examines ways to execute CDO computation using memristor crossbars, and work in [22,23] explores alternative algorithms specifically for the asset allocation task. However, none of our previous work has been implemented on the Loihi processor, or any other embedded system.

This paper is organized as follows: Section II provides a background overview of the general CECEP architecture, and Section III describes the *M* by *N* asset allocation problem. Section IV describes the presented spiking neuron solution to the asset allocation problem, and Section V discusses the Loihi implementation. Lastly, Section VI provides a results comparison and discussion, and Section VII provides a brief conclusion.

## II. THE CECEP ARCHITECTURE

The CECEP architecture is a net-centric execution framework for agents specified in a collection of agent-specification formalisms [7-10]. A more detailed description of this CECEP architecture can be found in [11].

### A. Cognitive Modeling

To enable rapid development, researchers are creating a domain-specific language (DSL) called the research modeling language (RML). In order to maximize scalability and interoperability during execution and simulation, RML requires users to conceive of and specify their models and agents as complex event processing agents [7].

The CECEP architecture incorporates model and agent capabilities based on *declarative*, *procedural*, and *domain* knowledge processing to the Esper framework. The result is an event-driven architecture that is capable of advanced cognitive modeling and complex event analytics. Research scientists can use this system to develop and field decision support, performance assessment, and instructional technologies based on cognitive models and agents.

### B. Cognitive Domain Ontologies

Cognitive Domain Ontologies (CDOs) formalize the CECEP Agent's domain knowledge. A CDO is a tree with alternating entities and relations. Entities correspond to domain objects, such as a playing card's value or suit. A CDO consists of three major entities: SubParts (SP), ChoicePoints (CP), and instances ([0..n]). A sample CDO is shown in Fig. 1.

SubParts contain a unification relationship with other entities, where all entities necessarily occur together (nodes Implication, and Evidence are SP nodes in Fig. 1). ChoicePoints represent an "either or" relationship between entities; only one may be active at a time. For example, the explanation ChoicePoint node in Fig. 1 can be either BrokenPipe, Raining, or NONE events. Finally, the instances relation captures replicated sub-structure. The CDO in Fig. 1 does not include an example of this type of structural relationship. Entities could have zero or more event attributes. User-defined constraint relationships allow users to connect events and attributes in a CDO to each other using conditionals (if, iff), connectives (and, or, not), and attribute comparisons (<, >, <=, >=, !=, etc.). Three constraints applicable to Implication events are listed in Table I.

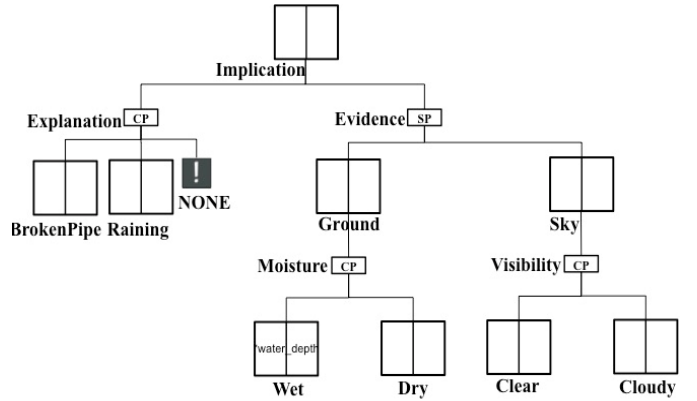

Fig. 1. CDO representing Implication events.

The combination of structural domain knowledge (Fig. 1) and relational domain knowledge (Table I) yield a complete CDO. CECEP agents process domain knowledge (CDOs) to deduce, abduces, reason, categorize, and plan. To enable CECEP agents to mine knowledge in CDOs, the event/relations networks underlying CDOs are translated into constraint networks and processed in a constraint solver.

Table I. Specifications of user-defined constraints relating explanation and evidence events.

| Name | Specification |
|---|---|
| **Raining** | IF Implication.explanation = Raining<br>THEN Implication.evidence.Ground.moisture = Wet<br>AND   Implication.evidence.Sky.visibility = Cloudy |
| **Broken Pipe** | IF Implication.explanation = BrokenPipe<br>THEN Implication.evidence.Ground.moisture = Wet<br>   OR<br>   (Implication.evidence.Sky.visibility = Clear<br>   AND<br>   NOT Implication.evidence.Ground.moisture = Dry) |
| **Dry Ground** | IFF NOT (Implication.explanation = Raining<br>OR<br>   Implication.explanation = BrokenPipe)<br>THEN Implication.evidence.Ground.moisture = Dry |
| **Wet Ground** | IFF Implication.evidence.Ground.moisture = Wet<br>THEN Implication.explanation = Raining<br>   OR<br>   Implication.explanation = BrokenPipe |

## III. THE M BY N ASSET ALLOCATION PROBLEM

The work in this paper discusses the implementation of a particular problem that is often tasked to the CDO: *M* by *N* asset allocation [12]. In this case we assume that $M$ tasks are present in an assignment description, and $N$ vehicles are available to resolve each task. Each task has a priority ($P$), a travel time relative to each vehicle ($TTA$), a probability of success ($S$), and a time to task completion ($TTC$). In this problem $TTC$ is equal to time to arrival ($TTA$) plus the time on task ($TOT$). It is the job of the CDO to determine the best possible way to allocate $N$ vehicles to $M$ tasks. The winning allocation decision is determined by the maximization of some objective function that produces some reward value based on a combination of these variables.

In this work, we implement the constraint that a given vehicle can only be assigned to one task. However, multiple vehicles can be assigned to the same task. An assignment of multiple vehicles to a single task may be the case if a given task has a high enough priority to justify this action.

One way to implement these decision making systems in hardware is to use GPUs to execute the instructions which generate the possible outcomes [12]. The problem with this approach is that the asset allocation problem can have an extremely large number of possible outcomes, and checking all of them with a GPU takes an unrealistic amount of time when optimized assignments are needed quickly. The number of possible outcomes for this problem can be calculated as $(N+1)^M$. For convenience, Table II shows the number of possible solutions for the type of asset allocation problems studied in this work.

Table II. Number of possible solutions for a set of allocation problems that vary in size.

| Allocation Problem Size | Number of Possible Solutions |
|---|---|
| 2 × 2 | 9 |
| 4 × 4 | 625 |
| 6 × 6 | 117,649 |
| 8 × 8 | 43,046,721 |
| 10 × 10 | 25,937,424,601 |

Table II shows that the number of possible solutions to this problem increases at a rapid exponential rate as problem size increases. With a problem size of 10 × 10 the number of possible solutions already exceeds 25 billion. Thus, the spiking algorithm presented in this work aids in dramatically reducing runtime when generating an optimal allocation solution.

IV. SPIKING NEURON IMPLEMENTATION

The proposed spiking neuron approach does not always predict the absolute best possible solution (that could be determined using an exhaustive search algorithm). However, it provides a near-optimal solution in a time that can be more than 1000× faster than the exhaustive search approach (depending on problem size). This system is based on a simple integrate and fire neuron model, where a single neuron is dedicated to each vehicle-target combination. The accumulation of each neuron is driven by a weight matrix that holds information relating to a specific allocation scenario. In this system, each neuron in the neuron grid is only required to fire a maximum of one time before the allocation result is determined. This is because winning vehicle-task combinations are determined by which neurons (each representing a possible vehicle-task combination) are able to fire first.

Fig. 2 displays a block diagram for the proposed spiking neuron approach. When in operation, each neuron in the system is subject to a constant uniform spiking input. This is due to the first layer of neurons being driven by a constant bias. Each neuron possesses a weight value that when combined with an input spike, produces a base accumulation rate (Fig. 2 (a)). Each neuron has a unique base accumulation rate that is determined by the set of parameters that correspond to a specific scenario.

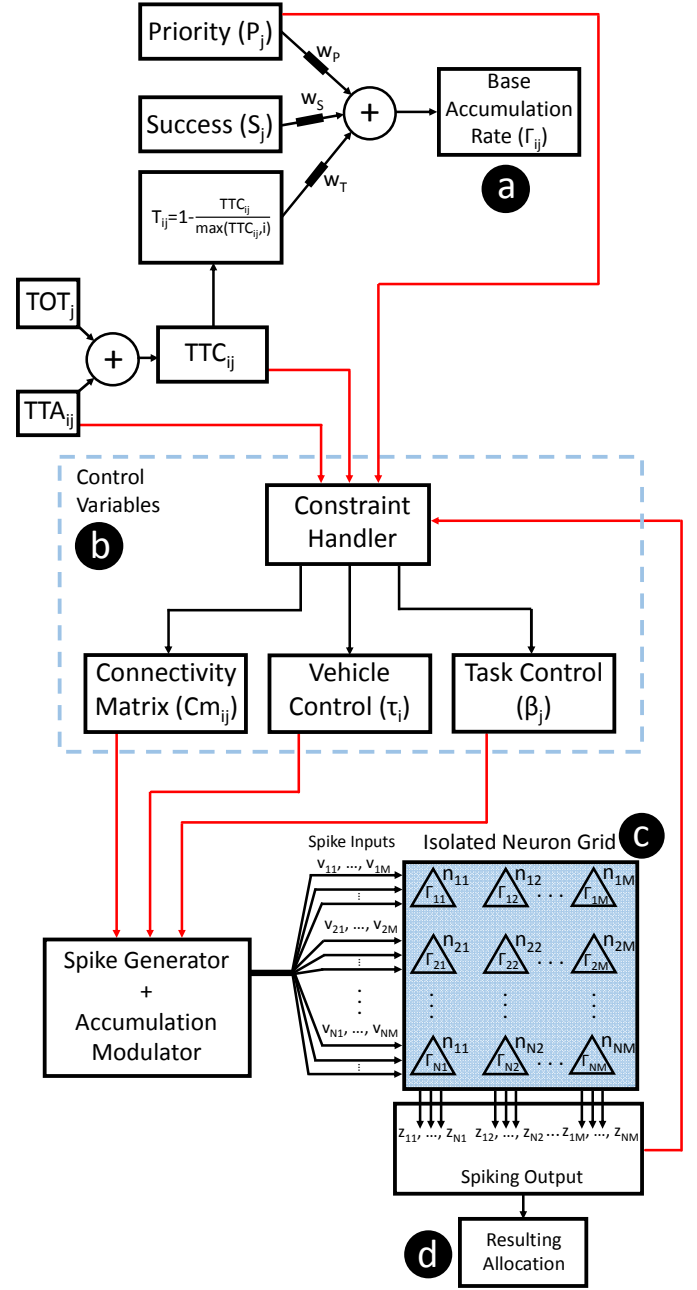

Fig. 2. Block diagram for the spiking neuron based asset allocation system showing the relationship between (a) the base accumulation rate, (b) the neuron control variables, (c) the isolated neuron grid, and (d) the resulting allocation output.

During operation, active neurons will accumulate until the winning set of neurons fires, resulting in a winning vehicle-task allocation combination. Whether or not a particular neuron is active depends of some additional control variables that are implemented within this system (see Fig. 2 (b)).

Each neuron ($n_{ij}$) in the isolated neuron grid (see Fig. 2 (c)) essentially accumulates according to equations (1) through (3). Each individual neuron has a unique base accumulation rate $\Gamma_{ij}$ as seen in equation (1) and in Fig. 2 (a). This rate is dependent on a weighted sum of task priority, task success, and relative task completion time. In equation (1), the weight values $w_P$, $w_S$, and $w_T$ correspond to priority ($P_j$), success ($S_j$) and completion time ($T_{ij}$) respectively. Each of the weights $w_P$, $w_S$, and $w_T$ are manually set to a value between 0 and 1. In this work, $w_P = 0.45$, $w_S = 0.1$, and $w_T = 0.5$.

$$\Gamma_{ij} = w_P P_j + w_S S_j + w_T T_{ij} \quad (1)$$

As shown in equation (2) the value $T_{ij}$ is determined by inverting the magnitude of the values in the $TTC_{ij}$ matrix. Thus, the biggest reward is placed upon the vehicle-task combinations that have the shortest completion times. In the denominator in equation (2), the maximum function collapses the $TTC_{ij}$ matrix along the $i$ dimension. This means that the denominator in equation (2) results in an array of size $M$ that holds the largest vehicle-task distance relative to each task. Thus, $T_{ij} = 0$ for all vehicle-task combinations with the longest $TTC$ for a given task. The value $T_{ij}$ becomes larger as $TTC$ is reduced.

$$T_{ij} = 1 - \frac{TTC_{ij}}{\max(TTC_{ij}, i)} \quad (2)$$

The base accumulation rate is then multiplied by a series of control variables to obtain the total accumulation rate $A_{ij}$ in equation (3).

$$A_{ij} = CM_{ij} \times \beta_j \times \tau_i \times \Gamma_{ij} \quad (3)$$

The parameter $CM_{ij}$ corresponds to the connectivity matrix. The connectivity matrix can be used to address constraints in more complex scenarios where only certain types of available vehicles are appropriate for specific types of tasks. Thus, $CM_{ij}$ could be set to 0 for any invalid vehicle-task combination to stop accumulation relating to incompatible pairs. Constraints could also be based on scenario parameters, which is why several parameters are shown as control inputs to the constraint handler.

The parameter $\beta_j$ is a dynamic variable can apply additional decay to certain neuron accumulation rates under certain circumstances. At the beginning of any cycle, $\beta_j$ is equal to $1/2^{D_j}$ for task $j$, where $D_j$ is equal to the number of vehicles already allocated to task $j$. This means that if a vehicle reaches task $j$, then all other vehicles will have a slowed accumulation in regards to task $j$. This increases the necessity of addressing a large number of tasks without eliminating the possibility that two vehicles can be allocated to the same task (because a high task priority may deem this necessary). At the beginning of a scenario, $D_j$ is 0 for all tasks.

The parameter $\tau_i$ is a binary variable that tracks which vehicles have been allocated to a task during the scenario runtime. If vehicle $i$ was successfully allocated, then all other neurons that correspond to vehicle $i$ should no longer accumulate. This is because a vehicle can only be allocated to one task within a scenario. Thus if $\tau_i = 0$ for each vehicle $i$, then the allocation problem is complete.

The parameters $\tau_i$ and $\beta_j$ are implemented in the Loihi system using a layer of feedback neurons. However, in [23] these parameters are implanted using logical operators that impact the spiking inputs. Therefore, in the general system block diagram in Fig. 2, the implementation of $\tau_i$ and $\beta_j$ is abstracted and handled by an accumulation modulator function that resides alongside the uniform spiking inputs. Details on the Loihi implementation will be discussed in the next section.

A visual representation for the isolated neuron grid corresponding to a 3 × 3 allocation problem is displayed in Fig. 3. One neuron is present for each vehicle-task combination. Therefore, a 3 × 3 allocation problem (where $M = N = 3$) requires 9 neurons. This diagram shows more specifically, the impact each parameter and control variable has on each neuron. Note that the $\beta$ values are constant across vehicle numbers, and the $\tau$ values are constant across task numbers.

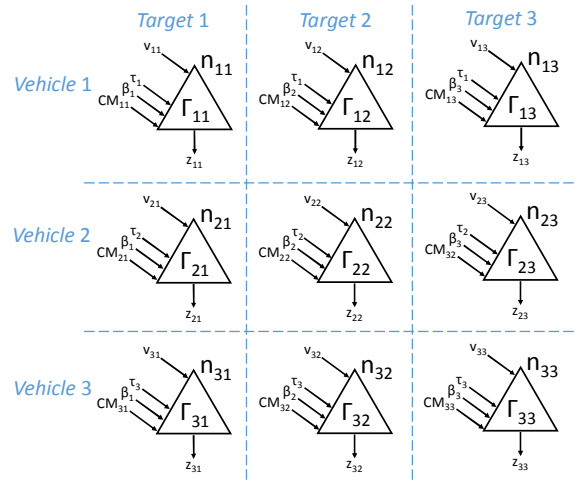

Fig. 3. Connection diagram for the proposed isolated neuron matrix.

## V. LOIHI IMPLEMENTATION

The Loihi processor was recently introduced by Intel [24,25]. The 60mm$^2$ chip was developed using Intel's 14nm process. Each chip contains approximately 1,000,000 neurons, and the system is capable of implementing hierarchical connectivity, dendritic compartments, synaptic delays, and programmable synaptic learning rules. Recently, it was shown that the Loihi system is approximately 38 times more power efficient for real-time DNN inference than a GPU based system [26]. For the following experiments, physical Loihi hardware utilized, which was accessed via remote login.

### A. Loihi System Setup and Programming

To execute this allocation algorithm using the Loihi system, the proposed algorithm was implemented using three layers of spiking neurons connected through weight matrices. The first neuron layer simply provides the input spikes, where the constant spike rate for each neuron is determined by an input bias. The second neuron layer contains the accumulating neurons. Each input neuron is connected to a unique accumulating neuron through a weight holding a value proportional to each neuron's base accumulation rate (see Fig. 2 (a)). The neurons in this layer will each accumulate until one is first to fire. Once a neuron in the accumulation layer fires, the third and final layer of neurons is able to implement feedback control. The control neuron layer is used to implemented the variables $\tau_i$ and $\beta_j$ within the Loihi system. In this system, for an $M$ by $N$ allocation, we require $M \times N$ input neurons, $M \times N$ accumulating neurons, and $M+N$ control neurons. The system requires an input and accumulation neuron for every vehicle-task combination. However, the system only requires one control neuron for each vehicle in addition to one control neuron for each task.

Diagrams for the neurons that exist in each of these three layers are displayed in Fig. 4. The input neurons in Fig. 4 (a) are simply driven by a bias input and produce a constant spiking output. The accumulation neurons in Fig. 4 (b) accumulate according to a base rate, but this rate can be slowed by the inhibitory *Task Control Input* or cancelled out by the inhibitory *Vehicle Control Input*. Finally, the control neurons as displayed in Fig. 4 (c) respond to an occurrence of a spike in the accumulation layer, and the control neurons possess recurrent feedback. The threshold of the control neurons is set in such a way that they will fire as a result of a single spike input. Thus, due to the recurrent feedback, a control neuron will fire every cycle after it has been told to fire once by an accumulation neuron.

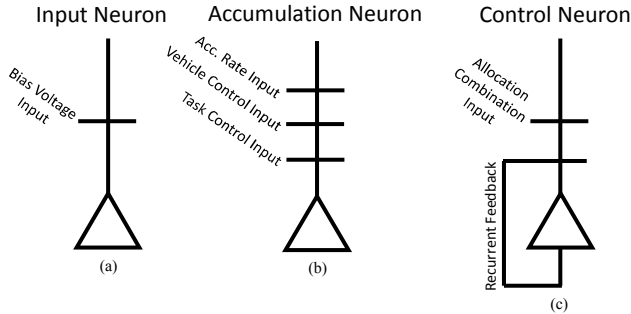

Fig. 4. Diagrams displaying basic neuron operation for (a) the input layer, (b) the accumulation layer, and (c) the control layer.

To implement the variable $\tau_i$, all accumulating neurons associated with vehicle $i$ must be set to no longer accumulate once vehicle $i$ reaches a target. This is done by connecting the control neuron associated with vehicle $i$ to the *Vehicle Control Input* of all accumulation neurons associated with vehicle $i$. The weight of this connection is set to –255, the absolute minimum value possible. Thus, accumulation neurons associated with vehicle $i$ will no longer fire because the excitatory accumulation rate will never exceed the inhibitory control connection.

Similarly, the variable $\beta_j$ is also implemented using neurons from the control layer. When an accumulation neuron fires that is associated with task $j$, all other accumulation neurons associated with task $j$ must reduce their accumulation rate by one half. Thus, the output of the control neuron associated with task $j$ is connected to the *Task Control Input* of all accumulation neurons associated with task $j$. The weight value for a task control connection is set to one fourth of the accumulation rate of a specific accumulation neuron in the inhibitory direction.

This achieves the desired result of slowing the accumulation rate by one half because once activated, the control neurons fire at twice the rate of the input neurons. This design decision was made to make sure control is enacted as quickly as possible after an accumulation neuron fires. The faster the control neurons fire, the less likely it will be that incorrect neurons will fire due to delay between the neuron layers.

The block diagram for the entire Loihi implementation of the spiking allocation algorithm is displayed in Fig. 5. The input bias voltage is applied to the layer of input neurons through a weight matrix that provides a one-to-one connection to the neurons in the accumulation layer. Whenever neuron $V_i T_j$ in the accumulation layer fires, an allocation designation is made so that Vehicle $i$ is allocated to Task $j$. Also, this output is applied to the control neuron layer through a splitting weight matrix so that the control neurons corresponding to Vehicle $i$ and Task $j$ are activated. The control neurons then provide the correct signals to implement the $\tau_i$ and $\beta_j$ variables in the algorithm.

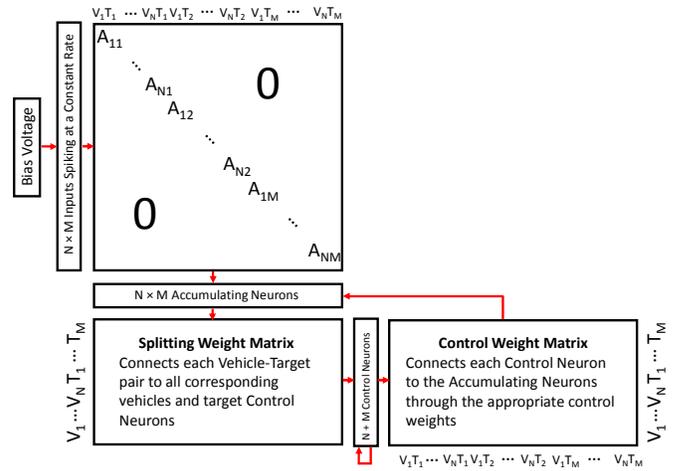

Fig. 5. Block diagram displaying connection diagram, weights, and neurons required to implement the M × N allocation algorithm on the Loihi system.

### B. Loihi Processor Execution Example

In this example a 4 × 4 allocation is executed using the Loihi system. Fig. 6 (a) shows the neuron voltage for each of the 16 accumulating neurons, and the four winning neurons display their accumulation functions with bold red outlines. Fig. 6 (b) shows the spiking pattern of all 8 control neurons. In Fig. 6 (a) the accumulation neuron with the highest accumulation rate fires just before the 80th tick mark. This neuron signifies that Vehicle 3 should be allocated to Task 1. Thus, in Fig. 6 (b) the control neurons corresponding to Vehicle 3 and Task 1 initiated spiking at this time. Due to this control neuron initiation, other vehicles associated with Task 1 will now have a reduced accumulation rate. For example, the accumulation neuron with the second highest accumulation rate also corresponds to Task 1, thus a clear accumulation rate change can be seen around the 80th tick mark. Likewise, the two neurons with the slowest accumulation rates are associated with Vehicle 3, so these two neuron accumulation values decay to 0 since Vehicle 3 is now assigned.

The spiking pattern associated with these accumulation neurons can be seen in Fig. 6 (c). This spiking pattern represents the resulting allocation. In this work the Vehicle-Task pairs are mapped to the neurons in a list fashion corresponding to the pattern $V_1T_1$, $V_1T_2$, …, $V_1T_M$, $V_2T_1$, $V_2T_2$, $V_2T_M$, …, $V_NT_1$, $V_NT_2$, …, $V_NT_M$. The result in Fig. 6 (c) shows that neurons 4, 5, 9, and 15 fired throughout the course of the allocation computation. These neurons correspond to the four cases: $V_1T_4$, $V_2T_1$, $V_3T_1$, and $V_4T_3$. This is the absolute best allocation that can be obtained for this scenario: [4 1 1 3]. Determination of allocation quality is discussed in the next section.

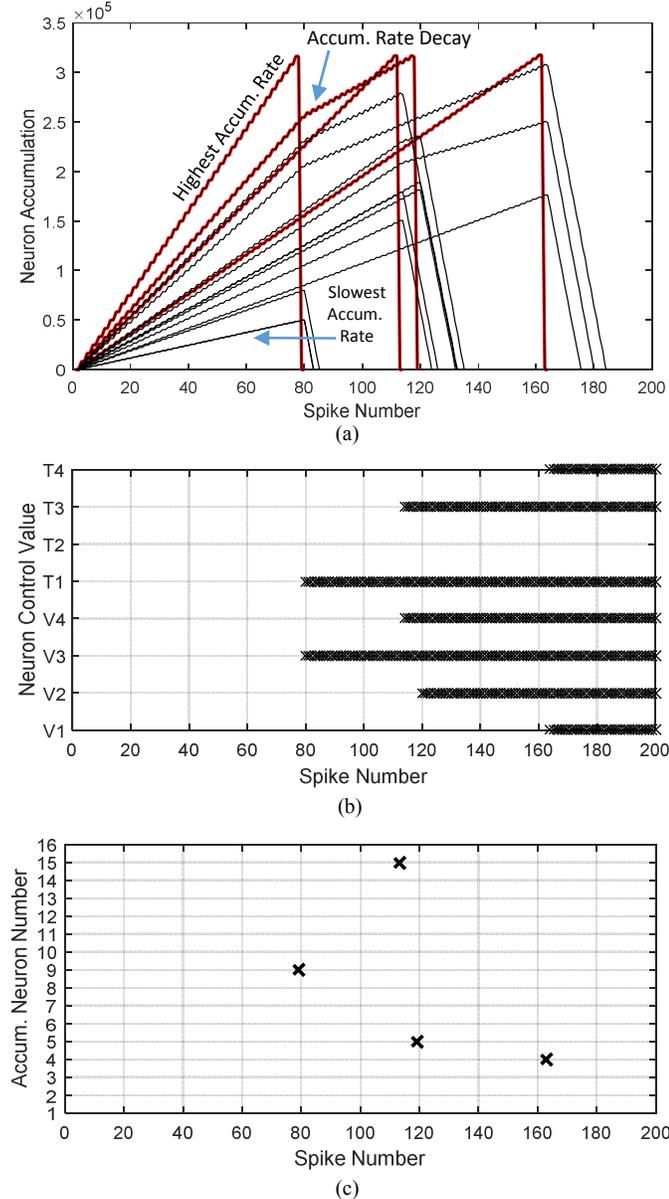

Fig. 6. Plots displaying the result of a 4 × 4 task allocation executed on the Loihi processor showing (a) the voltage values of all 16 accumulation neurons, (b) the spiking pattern of all 8 control neurons, and (c) the spiking pattern of the allocation neurons which signifies the allocation result.

## VI. RESULTS COMPARISON

Now that we have discussed the spiking neuron allocation system, the next step is to verify these solutions by determining if they match the optimal allocation combinations provided by the full scale GPU-based CDO solver. This is done using an exhaustive search to determine the best possible solution, in addition to all other solutions. This way, we can determine the precise rank of each solution provided by the spiking system by determining its position in a sorted solution list.

Table III displays baseline CDO reward, which represents the maximum possible optimization function output when considering the entire solution space, as well as the optimal allocation result when using the exhaustive search approach. In this table, the allocation result is listed so that array position corresponds to vehicle number and array value corresponds to task number. Table III also displays this information for the spiking system. However, the allocation results from the spiking system are also ranked according to how close they are to the optimal solution. This data was generated by sorting all possible solutions according to reward value. Thus, Table III shows that the answer generated by the spiking system is above the 99.9 percentile relative to the entire solution space in all cases. It should be noted that work in [23] presents a similar table, but the results generated by the exhaustive search approach are slightly different, even though the same scenarios were tested. This is because a point search method [22] was used in this work, and an area method was used in [23], meaning work in [23] was based on the optimization of a slightly different objective function.

The data in Table IV shows the runtime comparison for each of the executed scenarios between the exhaustive search and the proposed spiking algorithm. For the exhaustive search, a parallel search was performed on an NVidia Test K80 GPU. For the spiking algorithm, the runtimes on the Loihi processor were obtained using a runtime probe function (in software), which is available through the Loihi SDK. This table shows that as problem size increases, the speedup provided by the spiking system reaches two and three orders of magnitude. Note that the Loihi execution time is not perfectly monotonic as a function of problem size. This is because for the smaller allocation examples, execution time per time step levels off at around 2.4 µS, and execution time is then driven by number of time steps required for a given allocation. Given that accumulation neuron execution is performed in parallel, a greater number of accumulation neurons does not necessarily correlate to a larger accumulation time.

Table III. Summary of resulting allocations when using the Loihi spiking processor compared to the best possible solution generated by an exhaustive search.

| Allocation Size | Exhaustive Search Using GPU | | Spiking Neurons on Loihi System | | | |
|---|---|---|---|---|---|---|
| | Baseline CDO Reward | Baseline CDO Result | Effective Reward | Allocation Result | Answer Rank | Answer Percentile |
| 3×3 | 18.8703 | [2 1 1] | 18.8703 | [2 1 1] | 1 of 64 | 100% |
| 4×4 | 11.377 | [4 1 1 3] | 11.377 | [4 1 1 3] | 1 of 625 | 100% |
| 5×5 | 21.735 | [1 5 2 4 1] | 21.7223 | [1 5 2 1 1] | 3 of 7776 | 99.96% |
| 6×6 | 30.1986 | [2 4 1 5 3 4] | 28.0107 | [5 4 1 5 3 3] | 100 of 117649 | 99.91% |
| 7×7 | 47.8245 | [4 2 1 6 2 5 7] | 45.6475 | [4 2 4 6 5 4 7] | 166 of 2.09M | >99.99% |
| 8×8 | 40.8193 | [1 3 4 7 5 3 6 8] | 35.4964 | [1 3 3 3 5 5 4 7] | 7843 of 43.0M | 99.98% |

Table IV. Asset allocation runtime comparison between the exhaustive search algorithm running on an NVidia Tesla K80 GPU and the SNN algorithm running on the Loihi spiking processor for each simulated scenario.

| Allocation Size | CDO Search Time (GPU) | Loihi Execution Time | Loihi System Speedup |
|---|---|---|---|
| 3×3 | 224 ms | 0.312 ms | 717× |
| 4×4 | 231 ms | 0.384 ms | 601× |
| 5×5 | 233 ms | 0.319 ms | 730× |
| 6×6 | 234 ms | 0.414 ms | 565× |
| 7×7 | 269 ms | 0.428 ms | 629× |
| 8×8 | 955 ms | 0.737 ms | 1296× |

The most dramatic result in Table IV is that for an 8 × 8 allocation problem, the spiking system shows a speedup of more than 1200× while providing a solution in the 99.9 percentile. This dramatic increase in speed is due to the difference in computation amount in these two problems. The data in Table V shows that the 8 × 8 allocation can be performed using the Loihi system with only 144 neurons. This implies that it is likely possible to perform much larger allocation scenarios using the Loihi system. The reason we stopped at a maximum size of 8 × 8 in this work is because that was the largest size we were able to execute using the GPU based CDO solver at this time. Thus, that was the largest example that we could run to determine the optimal allocation. In the future we plan to run larger scale problems to determine power and timing requirements of these larger allocation examples.

Table V. Number of Loihi neurons required to perform problems allocation problems of different sizes.

| Allocation Problem Size | Number of Loihi Neurons |
|---|---|
| 2 × 2 | 12 |
| 3 × 3 | 24 |
| 4 × 4 | 40 |
| 5 × 5 | 60 |
| 6 × 6 | 84 |
| 7 × 7 | 112 |
| 8 × 8 | 144 |

However, as problem size increases in the Loihi system, the potential for limitation may be found in the maximum weight resolution. In the Loihi system, weights are limited to integer values between –255 and 255. If two competing neurons have similar accumulation values, there is a chance that they may fire at the same time incorrectly. For example, in some cases, the 8 × 8 allocation resulted in 9 neurons firing instead of 8, this led to two possible allocation combinations being recommended simultaneously. In this case, post-processing could possibly be used to choose the solution based on the which of the simultaneously firing neurons had a larger accumulation rate. This problem may also be remedied by using multiple neurons to represent one vehicle-target pair, which will be studied in future work. This would essentially increase the accumulation rate resolution of the system. This could also be improved by increasing the accumulation neuron firing threshold and increasing execution time to allow for the integration patterns to separate from each other.

## VII. CONCLUSION

This paper describes a spiking neuron approach for solving asset allocation problems in a faster, although more approximate way when compared to an exhaustive search. To execute the algorithm presented in this work, we used the Loihi Spiking Neural Network Processor to show that this approach can be implemented using low power specialized hardware. In all cases the Loihi system was able to generate an allocation solution with greater than 99.9% accuracy, with a speedup of more than 1000× in the most extreme case. In the future we plan to increase allocation problem size and study the potential limitations of allocation accuracy. We also plan to study the energy and timing trends as problem size is further increased.

ACKNOWLEDGMENT

This work was supported through funding from the Air Force Research Laboratory. Approved for public release: distribution is unlimited. 88ABW Cleared 03/20/19; 88ABW-2019-0845.